\documentclass{article}
\usepackage{ijcai17}
\usepackage[small]{caption}
\usepackage{amsfonts}
\usepackage{times}
\usepackage{amssymb}
\usepackage{amsmath}
\usepackage{graphicx}
\usepackage{subfig}
\usepackage[ruled,linesnumbered]{algorithm2e}
\usepackage{algorithmic}

\title{Deep Optical Flow Estimation Via Multi-Scale Correspondence Structure Learning}
\author{Shanshan Zhao$^1$, Xi Li$^{1,2,}$\thanks{Corresponding author}, Omar El Farouk Bourahla$^1$\\
$^1$ Zhejiang University, Hangzhou, China  \\
$^2$ Alibaba-Zhejiang University Joint Institute of Frontier Technologies, Hangzhou, China\\
\{zsszju,xilizju,obourahla\}@zju.edu.cn}

\begin{document}

\maketitle

\begin{abstract}
  As an important and challenging problem in computer vision, learning based optical flow estimation aims to discover the intrinsic correspondence structure between two adjacent video frames through statistical learning. Therefore, a key issue to solve in this area is how to effectively model the multi-scale correspondence structure properties in an adaptive end-to-end learning fashion. Motivated by this observation, we propose an end-to-end
  multi-scale correspondence structure learning (MSCSL) approach for optical flow estimation. In principle, the proposed MSCSL approach
  is capable of effectively capturing the multi-scale inter-image-correlation correspondence structures within a multi-level
  feature space from deep learning. Moreover, the proposed MSCSL approach builds a spatial Conv-GRU neural network model to adaptively
  model the intrinsic dependency relationships among these multi-scale correspondence structures. Finally, the above procedures for correspondence structure learning and multi-scale dependency modeling are implemented in a unified end-to-end deep learning framework.
  Experimental results on several benchmark datasets demonstrate the effectiveness of the proposed approach.

\end{abstract}

\section{Introduction}

Optical flow estimation seeks for perceiving the motion information across consecutive video frames, and has a wide range of vision applications
such as human action recognition and abnormal event detection. Despite the significant progress in the literature, optical flow estimation is still confronted with a number of difficulties in discriminative feature representation, correspondence structure modeling, computational flexibility, etc. In this paper, we focus on how to set up an effective learning pipeline that is capable of performing multi-scale correspondence structure modeling with discriminative feature representation in a flexible end-to-end deep learning framework.

Due to the effectiveness in statistical modeling, learning based approaches emerge as an effective tool of optical flow estimation~\cite{dosovitskiy2015flownet,jason2016back,ahmadi2016unsupervised,zhu2017guided}. Usually, these approaches either just take image matching at a single scale into account, or take a divide-and-conquer strategy that
copes with image matching at multiple scales layer by layer.
Under the circumstances of complicated situations (e.g., large inter-image displacement or complex motion), they are often
incapable of effectively capturing the interaction or dependency relationships among the multi-scale inter-image correspondence
structures, which play an important role in robust optical flow estimation. Furthermore, their matching strategies are often carried out
in the following two aspects. 1) Set a fixed range of correspondence at a single scale in the learning process~\cite{dosovitskiy2015flownet,jason2016back,zhu2017guided}; and 2) update the matching range dynamically with a coarse-to-fine scheme~\cite{ahmadi2016unsupervised,ranjan2016optical}. In practice, since videos have time-varying dynamic properties, selecting an appropriate fixed range for matching is difficult for adapting to various complicated situations. Besides, the coarse-to-fine scheme may cause  matching error propagations or accumulations from coarse scales to fine scales.
Therefore, for the sake of robust optical flow estimation, correspondence structure modeling ought to be performed in an adaptive multi-scale collaborative way. Moreover, it is crucial to effectively capture the cross-scale dependency information while preserving spatial self-correlations for each individual scale in a totally data-driven fashion.

\begin{figure*}[th]
\centering
\includegraphics[scale=0.5]{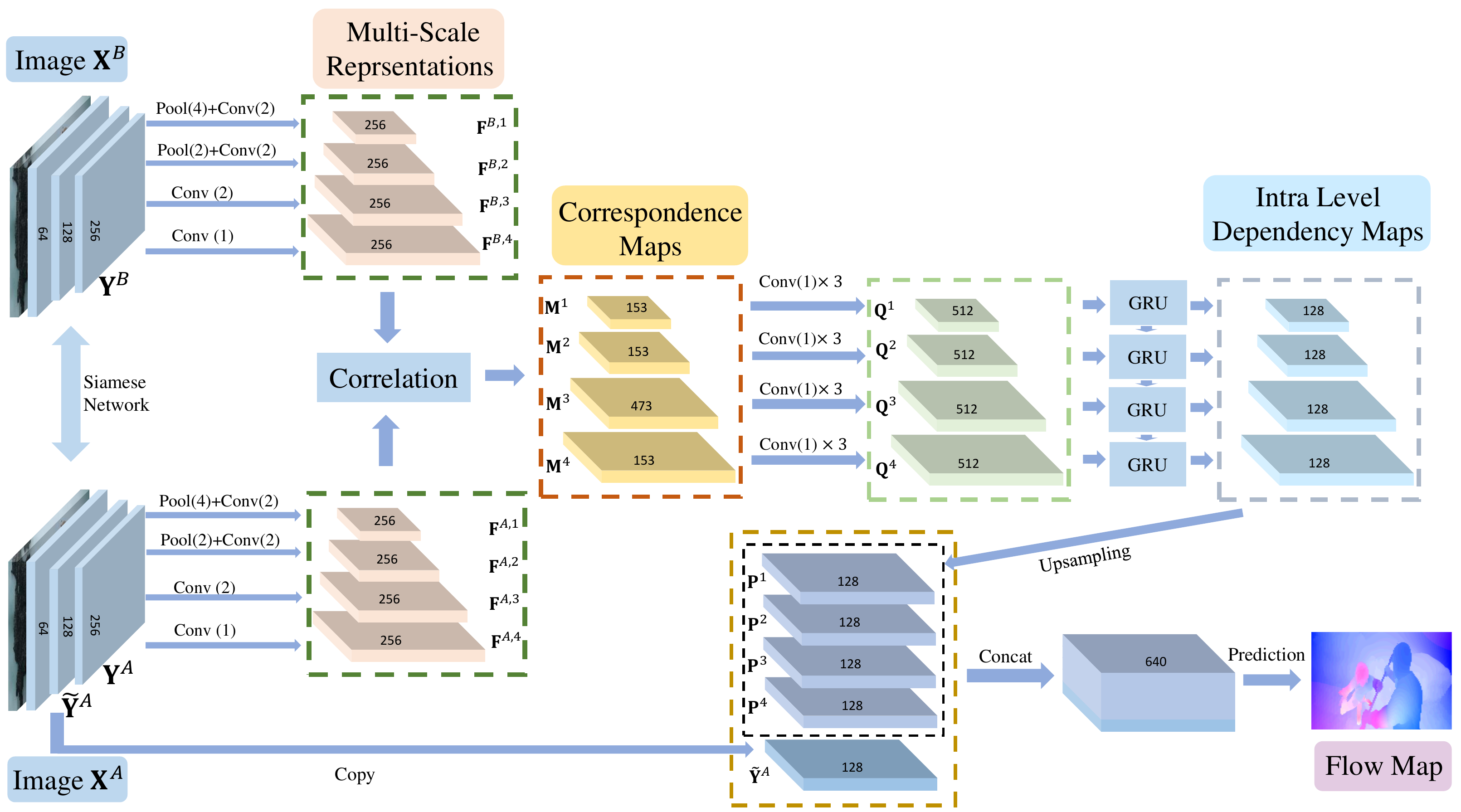}
\caption{The proposed CNN framework of Multi-Scale Correspondence Structure Learning (MSCSL). The $n$ in Pool($n$) and Conv($n$) denotes the $stride$ of corresponding operation, and $\times{3}$ denotes three consecutive operations. The network consists of three parts: (1) Multi-Scale Correspondence Structure Modelling, this part uses a Siamese Network to extract robust multi-level deep features for the two images, and then constructs the correspondence structures between the feature maps at different scales, (2) Correspondence Maps Encoding, this part employs a Spatial Conv-GRU presented in this work to encode the correspondence maps at different scales, (3) Prediction, we use the encoded feature representation to predict the optical flow map.}
\label{architecture}
\end{figure*}

Motivated by the above observations, we propose a novel unified end-to-end optical flow estimation approach called \textbf{Multi-Scale Correspondence Structure Learning (MSCSL)} (as shown in Fig.~\ref{architecture}), which jointly models the dependency of multi-scale correspondence structures by a Spatial Conv-GRU neural network model based on multi-level deep learning features.
To summarize, the contributions of this work are twofold:
\begin{itemize}
\item We propose a multi-scale correspondence structure learning approach, which captures the multi-scale inter-image-correlation correspondence structures based on the multi-level deep learning features. As a result, the task of optical flow estimation is accomplished by jointly learning the inter-image correspondence structures at multiple scales within an end-to-end deep learning framework. Such a multi-scale correspondence structure learning approach is innovative in optical flow estimation to the best of our knowledge.

\item We design a Spatial Conv-GRU neural network model to model the cross-scale dependency relationships among the multi-scale correspondence structures while preserving spatial self-correlations for each individual scale in a totally data-driven manner. As a result, adaptive multi-scale matching information fusion is enabled to make optical flow estimation adapt to various complicated situations, resulting in robust estimation results.
\end{itemize}

\section{Our Approach}

\subsection{Problem Formulation}
Let $\{(\mathbf{X}_i,\mathbf{O}_i)\}_{i=1}^N$ be a set of $N$ training samples, where $\mathbf{X}_i=(\mathbf{X}_i^A\in{\mathbb{R}^{H\times{W}\times{3}}}, \mathbf{X}_i^B\in{\mathbb{R}^{H\times{W}\times{3}}})$ and $\mathbf{O}_i\in{\mathbb{R}^{H\times{W}\times{2}}}$ represent an RGB image pair and the corresponding optical flow respectively. In this paper, our objective is to learn a model $f(\mathbf{X}_i;\mathbf{\theta})$ parameterized by $\mathbf{\theta}$ to predict the dense motion of the first image $\mathbf{X}_i^A$. For the sake of expression, we ignore the left subscript $i$ in the remaining parts.

In this paper, we focus on two factors, (1) computing the correlation maps between image representations at different scales and adaptively setting up the correspondence structure in a data-driven way, (2) encoding the correspondence maps into high-level feature representation for regressing the optical flow.

\subsection{Multi-Scale Correspondence Structure Modelling}
\textbf{Multi-Scale Image Representations.}
To represent the input image at multiple scales, we firstly employ convolution neural networks (CNNs) to extract the deep features at a single scale parameterized by $\mathbf{\theta}_1$ to represent the image $\mathbf{I}$, as illustrated in Fig.~\ref{architecture}:
\begin{equation}
\label{sr}
\mathbf{Y}=f_{CNN1}(\mathbf{I};\mathbf{\theta}_1)
\end{equation}
and then model the multi-level feature representations parameterized by $\{\mathbf{\theta}_{2,l}\}$ with $\mathbf{Y}$ as the input, as depicted in Fig.~\ref{architecture}:
\begin{equation}
\label{representation}
\mathbf{F}^l=f_{CNN2}(\mathbf{Y};\mathbf{\theta}_{2,l},l=1,2,\dots)
\end{equation}
where $\mathbf{F}^l$ represents the $l$-th level, and the size of $\mathbf{F}^{l+1}$ is larger than that of $\mathbf{F}^l$. From top to bottom (or coarse to fine), the feature representations at small scales\footnote{In this paper, the small scale means small size; the large scale means large size} tend to learn the sematic components, which contribute to find the correspondence of semantic parts with large displacements; Furthermore, the large scale feature maps tend to learn the local representation, which can distinguish the patches with small displacements.
In this paper, we use $\{\mathbf{F}^{A,l}\}$ and $\{\mathbf{F}^{B,l}\}$ to denote the multi-scale representations of $\mathbf{X}^A$ and $\mathbf{X}^B$ respectively.

\noindent\textbf{Correspondence Structure Modelling.} Given an image pair $(\mathbf{X}^A,\mathbf{X}^B)$ from a video sequence, we firstly extract their multi-level feature representations $\{\mathbf{F}^{A,l}\}$ and $\{\mathbf{F}^{B,l}\}$ using Eq.~\ref{sr} and Eq.~\ref{representation}. In order to learn the correspondence structures between the image pair, we calculate the similarity between the corresponding feature representations instead. Firstly, we discuss the correlation computation proposed in~\cite{dosovitskiy2015flownet}:
\begin{equation}
\label{correlation}
\begin{split}
Corr(\mathbf{F}_{i,j}^A,&\mathbf{F}_{S(i,j;d)}^B)=\\
&Concat\{\sum_{o_x=-k}^k\sum_{o_y=-k}^k\langle{
\mathbf{F}_{i+o_x,j+o_y}^A,\mathbf{F}_{p+o_x,q+o_y}^B
}\rangle,\\
& (p,q)\in{[i-d,i+d]\times{[j-d,j+d]}}\}
\end{split}
\end{equation}
where $\mathbf{F}_{i,j}^A$ and $\mathbf{F}_{i,j}^B$ denote the feature vector at the $(i,j)$-th location of $\mathbf{F}^A$ and $\mathbf{F}^B$ respectively, and $Concat\{\cdot\}$ denotes concatenating the elements in the set $\{\cdot\}$ to a vector, $S(i,j;d)$ denotes the $(2d+1)\times{(2d+1)}$ neighborhood of location $(i,j)$. The meaning of Eq.~\ref{correlation} is that given a maximum displacement $d$, the correlations between the location $(i,j)$ in $\mathbf{F}^A$ and $S(i,j;d)$ in $\mathbf{F}^B$ can be obtained by computing the similarities between the square patch of size $(2k+1)\times{(2k+1)}$ centered at location $(i,j)$ in $\mathbf{F}^A$ and square patches of the same size centered at all locations of $S$ in $\mathbf{F}^B$.

To model the correspondence between the $(i,j)$-th location in $\mathbf{F}^A$ and its corresponding location $(\hat{i},\hat{j})$ in $\mathbf{F}^B$, we can (1) calculate $Corr(\mathbf{F}_{i,j}^A,\mathbf{F}_{S(i,j;d)}^B)$ in a small neighbourhood $S$ of the $(i,j)$-th location in $\mathbf{F}^B$, or (2) calculate $Corr(\mathbf{F}_{i,j}^A,\mathbf{F}_{S(i,j;d)}^B)$ in a large enough neighbourhood $S$ of the $(i,j)$-th location in $\mathbf{F}^B$, or even in the whole feature map $\mathbf{F}^B$. But the former can not guarantee the computation of similarity between the $(i,j)$-th location and the corresponding $(\hat{i},\hat{j})$-th location, while the latter leads to low computational efficiency, because the complexity $\mathcal{O}(d^2k^2)$ of Eq.~\ref{correlation} exhibits quadratic growth when the value of $d$ increases.
To address that problem, we adopt correlation computation at each scale of multi-scale feature representations $\{\mathbf{F}^{A,l}\}$ and $\{\mathbf{F}^{B,l}\}$:
\begin{equation}
\label{cmap}
\mathbf{M}_{i,j}^l=Corr(\mathbf{F}_{i,j}^{A,l},\mathbf{F}_{S(i,j;d_l)}^{B,l})
\end{equation}
where the maximum displacement $d_l$ varies from bottom to top.

In order to give the network more flexibility in how to deal with the correspondence maps, we add three convolutional layers to the outputs of the $Correlation$ operation, which is the same as that proposed in~\cite{dosovitskiy2015flownet}, to extract the high-level representations parameterized by $\{\mathbf{\theta}_{3,l}\}$, as described in Fig.~\ref{architecture}:
\begin{equation}
\label{rmaps}
\mathbf{Q}^l=f_{CNN3}(\mathbf{M}^l;\mathbf{\theta}_{3,l}, l=1,2,\dots)
\end{equation}

\subsection{Correspondence Maps Encoding Using Spatial Conv-GRU}
\noindent\textbf{Cross-Scale Dependency Relationships Modelling.} For the sake of combining the correlation representations $\{\mathbf{Q}^l\}$ and preserving the spatial structure to estimate dense optical flow, we consider the representations as a feature map sequence, and then apply Convolutional Gated-Recurrent-Unit Recurrent Networks(Conv-GRUs) to model the cross-scale dependency relationships among the multi-scale correspondence structures. Conv-GRUs have been used to model the temporal dependencies between frames of the video sequence~\cite{ballas2015delving,siam2016convolutional}. A key advantage of Conv-GRUs is that they can not only model the dependencies among a sequence, but also preserve the spatial location of each feature vector. One of significant differences between a Conv-GRU and a traditional GRU is that innerproduct operations are replaced by convolution operations.

However, because of the employed scheme similar to coarse-to-fine, the size of the $(t+1)$-th input in the sequence is larger than that of the $t$-th input. We cannot apply the standard Conv-GRU on our problem, so instead we propose a Spatial Conv-GRU in which each layer's output is upsampled as the input of the next layer. For the input sequence $\{\mathbf{Q}^l\}$, the formulation of the Spatial Conv-GRU is:
\begin{align}
\label{scgru}
&\mathbf{Z}^l=\sigma(\mathbf{W}_z*\mathbf{Q}^l+\mathbf{U}_z*\mathbf{H}^{l-1,\uparrow})\\
&\mathbf{R}^l=\sigma(\mathbf{W}_r*\mathbf{Q}^l+\mathbf{U}_r*\mathbf{H}^{l-1,\uparrow})\\
&\widetilde{\mathbf{H}}^l=tanh(\mathbf{W}*\mathbf{Q}^l+\mathbf{U}*(\mathbf{R}^l\odot{\mathbf{H}^{l-1,\uparrow}}))\\
&\mathbf{H}^l=(\mathbf{1}-\mathbf{Z}^l)\odot{\mathbf{H}^{l-1,\uparrow}}+\mathbf{Z}^l\odot{\widetilde{\mathbf{H}}^l}\\
&\mathbf{H}^{l,\uparrow}=\mathbf{W}^{\uparrow}\circledast\mathbf{H}^l
\end{align}
where $*$ and $\odot$ denote a convolution operation and an element-wise multiplication respectively, and $\sigma$ is an activation function, e.g., $sigmoid$., $\circledast$ denotes the transposed convolution. The Spatial Conv-GRU can model the transition from coarse to fine and recover the spatial topology, outputting intra-level dependency maps $\{\mathbf{H}^l\}$.

\begin{table*}[htp]\tiny
\caption{Comparison of average endpoint errors (EPE) to the state-of-the-art. The times with right superscript $*$ indicate that the methods run on CPU, while the rest run on GPU. The numbers in parentheses are the results of the networks on dataset they were fine-tuned on. And the methods with +ft represent that the models were fine-tuned on MPI Sintel training dataset (two versions together) after trained on Flying Chairs training dataset.}
\label{epe}
\centering
\begin{tabular}[t]{|p{70pt}|cc|cc|cc|c|c|c|}
\hline
Methods & \multicolumn{2}{c|}{Sintel clean} & \multicolumn{2}{c|}{Sintel final} & \multicolumn{2}{c|}{KITTI 2012} & Middlebury & Flying Chairs & Time (sec) \\
& train & test & train & test & train & test & train & test & \\
\hline
EpicFlow & $2.40$ & $4.12$ & $3.70$ & $6.29$ & $3.47$ & $3.80$ & $0.31$ & $2.94$ & $16^*$ \\
DeepFlow & $3.31$ & $5.38$ & $4.56$ & $7.21$ & $4.58$ & $5.80$ & $0.21$ & $3.53$ & $17^*$ \\
FlowFields & $1.86$ & $3.75$ & $3.06$ & $5.81$ & $3.33$ & $3.50$ & $8.33$ & $0.27$ & $22^*$ \\
EPPM & $-$ & $6.49$ & $-$ & $8.38$ & $-$ & $9.20$ & $-$ & $-$ & $0.2$ \\
DenseFlow & $-$ & $4.39$ & $-$ & $7.42$ & $-$ & $2.90$ & $-$ & $-$ & $~265^*$ \\
LDOF & $4.64$ & $7.56$ & $5.96$ & $9.12$ &$10.94$ & $12.40$ & $0.44$ & $3.47$ & $65^*$ \\
\hline
FlowNetS & $4.50$ & $7.42$ & $5.45$ & $\mathbf{8.43}$ & $8.26$ & $-$ & $1.09$ & $2.71$ & $0.08$ \\
FlowNetC & $4.31$ & $7.28$ & $5.87$ & $8.81$ &$9.35$ & $-$ & $1.15$ & $2.19$ & $0.15$ \\
SPyNet & $4.12$ & $\mathbf{6.69}$ & $5.57$ & $\mathbf{8.43}$ & $9.12$ & $-$ & $\mathbf{0.33}$ & $2.63$ & $0.07$ \\
MSCSL/wosr & $3.63$ & $-$ & $4.93$ & $-$ & $5.98$ & $-$ & $0.87$ & $2.14$ & $\mathbf{0.05}$ \\
MSCSL/wor & $\mathbf{3.37}$ & $-$ & $4.72$ & $-$ & $\mathbf{5.80}$ & $-$ & $0.92$ & $2.11$ & $0.06$ \\
MSCSL & $3.39$ & $-$ & $\mathbf{4.70}$ & $-$ & $5.87$ & $-$ & $0.90$ & $\mathbf{2.08}$ & $0.06$ \\
\hline
FlowNetS+ft & $(3.66)$ & $6.97$ & $(4.44)$ & $7.76$ & $7.52$ & $9.10$ & $0.98$ & $3.04$ & $0.08$ \\
FlowNetC+ft & $(3.78)$ & $6.85$ & $(5.28)$ & $8.51$ &$8.79$ & $-$ & $0.93$ & $2.27$ & $0.15$ \\
SPyNet+ft & $(3.17)$ & $6.64$ & $(4.32)$ & $8.36$ & $8.25$ & $10.10$ & $\mathbf{0.33}$ & $3.07$ & $0.07$ \\
MSCSL/wosr+ft & $(3.18)$ & $\mathbf{5.68}$ & $(4.21)$ & $7.49$ & $5.89$ & $6.90$ & $0.81$ & $2.51$ & $\mathbf{0.05}$ \\
MSCSL/wor+ft & $(\mathbf{3.07})$ & $5.79$ & $(4.16)$ & $\mathbf{7.42}$ & $5.87$ & $\mathbf{6.80}$ & $0.87$ & $2.28$ & $0.06$ \\
MSCSL+ft & $(\mathbf{3.07})$ & $5.78$ & $(\mathbf{4.15})$ & $\mathbf{7.42}$ & $\mathbf{5.77}$ & $7.10$ & $0.86$ & $\mathbf{2.25}$ & $0.06$ \\
\hline
\end{tabular}
\end{table*}

\noindent\textbf{Intra-Level Dependency Maps Combination.} After getting the hidden outputs $\{\mathbf{H}^l\}$, we upsample them to the same size, written as $\mathbf{P}^l$:
\begin{equation}
\label{fuse}
\mathbf{P}^l=f_{CNN4}(\mathbf{Q}^l;\mathbf{\theta}_4)
\end{equation}
where $\mathbf{\theta}_4:=\{\mathbf{W}_z,\mathbf{U}_z,\mathbf{W}_r,\mathbf{U}_r,\mathbf{W},\mathbf{U},\mathbf{W}^\uparrow\}$ are the parameters needed to be optimized. Furthermore, we concatenate the hidden outputs $\{\mathbf{P}^l\}$ with the $2$nd convolutional output $\widetilde{\mathbf{Y}}^A$ of $\mathbf{X}^A$ to get the final encoded feature representation for optical flow estimation, as depicted in Fig.~\ref{architecture}:
\begin{equation}
\label{concat}
\mathbf{E}=Concat\{\widetilde{\mathbf{Y}}^A,\mathbf{P}^l,l=1,2,\dots\}
\end{equation}
where $Concat$ represents the concatenation operation.

Finally, the proposed framework learns a function parameterized by $\mathbf{\theta}_5$ to predict the optical flow:
\begin{equation}
\label{predict}
\begin{split}
\hat{\mathbf{O}}&=f_{CNN5}(\mathbf{E};\mathbf{\theta}_5)\\
&=f(\mathbf{X}^A,\mathbf{X}^B;\theta_1,\theta_{2,l},\theta_{3,l},\theta_4,\theta_5,l=1,2,\dots)\\
\end{split}
\end{equation}
\subsection{Unified End-to-End Optimization}\label{lf}
As the image representation, correspondence structure learning and correspondence maps encoding are highly related, we construct a unified end-to-end framework to optimize the three parts jointly.
The loss function used in the optimization framework consists of two parts, namely, a supervised loss and an unsupervised loss (or reconstruction loss). The former is the endpoint error (EPE), which measures the Euclidean distance between the predicted flow $\hat{\mathbf{O}}$ and the ground truth $\mathbf{O}$, while the latter is based on the brightness constancy assumption, which measures the Euclidean distance between the first image $\mathbf{X}^A$ and the warped second image $\mathbf{X}^{B_{warp}}$.
\begin{align}
\label{loss}
&\mathcal{L}(\mathbf{O},\hat{\mathbf{O}};\mathbf{X}^A,\mathbf{X}^B)=\mathcal{L}_s(\mathbf{O},\hat{\mathbf{O}})+\lambda\mathcal{L}_{us}(\hat{\mathbf{O}};\mathbf{X}^A,\mathbf{X}^B)\\
\label{loss_s}
&\mathcal{L}_s(\mathbf{O},\hat{\mathbf{O}})=\sum_{i,j}\sqrt{(\mathbf{O}_{i,j}^u-\hat{\mathbf{O}}_{i,j}^u)^2+(\mathbf{O}_{i,j}^v-\hat{\mathbf{O}}_{i,j}^v)^2}\\
\label{loss_us}
&\mathcal{L}_{us}(\hat{\mathbf{O}};\mathbf{X}^A,\mathbf{X}^B)=\sum_{i,j}\sqrt{(\mathbf{X}_{i,j}^A-\mathbf{X}_{i,j}^{B_{warp}})^2}
\end{align}
where $\hat{\mathbf{O}}^u$ and $\hat{\mathbf{O}}^v$ denote the displacement in horizontal and vertical respectively, and $\lambda$ is the balance parameter. $\mathbf{X}^{B_{warp}}$ can be calculated via bilinear sampling according to $\hat{\mathbf{O}}$, as proposed in Spatial Transform Networks\cite{jaderberg2015spatial}:
\begin{equation}
\label{warp}
\begin{split}
\mathbf{X}^{B_{warp}}_{i,j}=\sum_n^H\sum_m^W\mathbf{X}^B_{m,n}
&max(0,1-\|i+\hat{\mathbf{O}}_{i,j}^u-m\|)\\
&max(0,1-\|j+\hat{\mathbf{O}}_{i,j}^v-n\|)
\end{split}
\end{equation}

\begin{table*}[htp]
\tiny
\vspace{-1em}
\caption{Comparison of FlowNet, SPyNet and our proposed methods on MPI Sintel test datasets for different velocities ($s_*$) and displacement ($d_*$).}
\label{epesintel}
\centering
\begin{tabular}[t]{|p{50pt}|cccccc|}
\hline
Methods & \multicolumn{6}{c|}{Sintel Final}  \\
& $d_{0-10}$ & $d_{10-60}$ & $d_{60-140}$ & $s_{0-10}$ & $s_{10-40}$ & $s_{40+}$  \\
\hline
FlowNetS+ft & $7.25$ & $4.61$ & $2.99$ & $1.87$ & $5.83$ & $43.24$  \\
FlowNetC+ft & $7.19$ & $4.62$ & $3.30$ & $2.30$ & $6.17$ & $\mathbf{40.78}$  \\
SPyNet+ft & $6.69$ & $4.37$ & $3.29$ & $\mathbf{1.39}$ & $5.53$ & $49.71$  \\
MSCSL/wosr+ft & $6.27$ & $3.77$ & $2.96$ & $1.96$ & $4.97$ & $40.98$ \\
MSCSL/wor+ft & $6.08$ & $\mathbf{3.57}$ & $\mathbf{2.79}$ & $1.76$ & $\mathbf{4.81}$ & $41.74$  \\
MSCSL+ft & $\mathbf{6.06}$ & $3.58$ & $2.81$ &  $1.73$ & $4.83$ & $41.87$ \\
\hline
Methods & \multicolumn{6}{c|}{Sintel Clean} \\
& $d_{0-10}$ & $d_{10-60}$ & $d_{60-140}$ & $s_{0-10}$ & $s_{10-40}$ & $s_{40+}$ \\
\hline
FlowNetS+ft &  $5.99$ & $3.56$ & $2.19$ & $1.42$ & $3.81$ & $40.10$ \\
FlowNetC+ft &  $5.57$ & $3.18$ & $1.99$ & $1.62$ & $3.97$ & $\mathbf{33.37}$ \\
SPyNet+ft &  $5.50$ & $3.12$ & $1.71$ & $\mathbf{0.83}$ & $3.34$ & $43.44$ \\
MSCSL/wosr+ft &  $4.84$ & $2.39$ & $1.64$ & $1.27$ & $3.26$ & $33.40$\\
MSCSL/wor+ft  & $4.80$ & $2.34$ & $1.61$ & $1.26$ & $\mathbf{3.07}$ & $34.90$ \\
MSCSL+ft & $\mathbf{4.79}$ & $\mathbf{2.33}$ & $\mathbf{1.58}$ & $1.24$ & $3.08$ & $34.83$\\
\hline
\end{tabular}
\end{table*}
Because the raw data $\mathbf{X}^A$ and $\mathbf{X}^B$ contain noise and illumination changes and are less discriminative, in some cases the brightness constancy assumption is not satisfied; Furthermore, in highly saturated or very dark regions, the assumption also suffers difficulties~\cite{jason2016back}. Therefore, applying Eq. \ref{loss_us} on the raw data directly will make the network more difficult when training. To address that issue, we apply the brightness constancy assumption on the $2$nd convolutional outputs $\widetilde{\mathbf{Y}}^A$ and $\widetilde{\mathbf{Y}}^B$ of $\mathbf{X}^A$ and $\mathbf{X}^B$ instead of $\mathbf{X}^A$ and $\mathbf{X}^B$. The training and test stages are shown in Alg.~\ref{alg:train-test}.
\begin{algorithm}[t]\footnotesize
\caption{Deep Optical Flow Estimation Via MSCSL}
\label{alg:train-test}
    \KwIn{ A set of $N$ training samples $\{((\mathbf{X}_i^A,\mathbf{X}_i^B),\mathbf{O}_i)\}_{i=1}^N$}
    \KwOut{ The deep model parameterized by $\theta$: $f(\mathbf{X}^A,\mathbf{X}^B;\theta)$}
\tcc{The training stage}
    \Repeat{$iter<$ max\_iter}
    {
        \tcc{For the $K$ batches, do}
        \For{$k=1,\dots,K$}
        {
         \tcc{Process the $k$-th training mini-batches $\mathcal{B}_k$}
         \For{$n\in{\mathcal{B}_k}$}
         {
         \tcc{Process the $n$-th image pair in $\mathcal{B}_k$}
         Extract the image representation $\mathbf{Y}_n^A$ and $\mathbf{Y}_n^B$ using Eq.~\ref{sr}\;
         Model the multi-scale feature representation $\{\mathbf{F}_n^{A,l}\}$ and $\{\mathbf{F}_n^{B,l}\}$ using Eq.~\ref{representation}\;
         Compute the correlation between feature representations $\{\mathbf{M}_n^l\}$ using Eq.~\ref{correlation} and Eq.~\ref{cmap}\;
         Extract the high-level representations $\{\mathbf{Q}_n^l\}$ of $\{\mathbf{M}_n^l\}$ using Eq.~\ref{rmaps}\;
         Encode the correspondence representations $\{\mathbf{Q}_n^l\}$ to get $\{\mathbf{P}_n^l\}$ using Eq.~\ref{scgru}\;
         Concatenate $\{\mathbf{P}_n^l\}$ with the $2$nd convolutional outputs of $\mathbf{X}_n^A$ to obtain $\mathbf{E}_n$ using Eq.~\ref{concat}\;
         Regress the optical flow estimation $\hat{\mathbf{O}}_n$ using Eq.~\ref{predict}\;
         Minimize the objective function Eq.~\ref{loss}\;
         }
         \tcc{Update network parameters}
         Update parameters $\theta=\{\theta_1,\theta_{2,l},\theta_{3,l},\theta_4,\theta_5,l=1,2,\dots\}$ using Adam;
        }
        $iter\leftarrow{iter+1}$
    }
\Return\;
\tcc{The test stage}
\KwIn{Given an image pair $(\mathbf{X}^{A_t},\mathbf{X}^{B_t})$ and the trained deep model $f(\mathbf{X}^A,\mathbf{X}^B;\theta)$}
\KwOut{The predicted optical flow $\hat{\mathbf{O}}^t=f(\mathbf{X}^{A_t},\mathbf{X}^{B_t};\theta_1,\theta_{2,l},\theta_{3,l},\theta_4,\theta_5,l=1,2,\dots)$}
    \Return \;
\end{algorithm}
\section{Experiments}

\subsection{Datasets}
\textbf{Flying Chairs}~\cite{dosovitskiy2015flownet} is a synthetic dataset created by applying affine transformations to a real image dataset and a rendered set of 3D chair models. This dataset contains $22,872$ image pairs, and is split into $22,232$ training and $640$ test pairs.

\noindent\textbf{MPI Sintel}~\cite{butler2012naturalistic} is created from an animated movie and contains many large displacements and provides dense ground truth. It consists of two versions: the Final version and the Clean version. The former contains motion blurs and atmospheric effects, while the latter does not include these effects. There are $1,041$ training image pairs for each version.

\noindent\textbf{KITTI 2012}~\cite{geiger2012we} is created from real world scenes by using a camera and a 3D laser scanner. It consists of $194$ training image pairs with sparse optical flow ground truth.

\noindent\textbf{Middlebury}~\cite{baker2011database} is a very small dataset, containing only $8$ image pairs for training. And the displacements are typically limited to $10$ pixels.

\begin{figure}[t]
\centering
\includegraphics[scale=0.26]{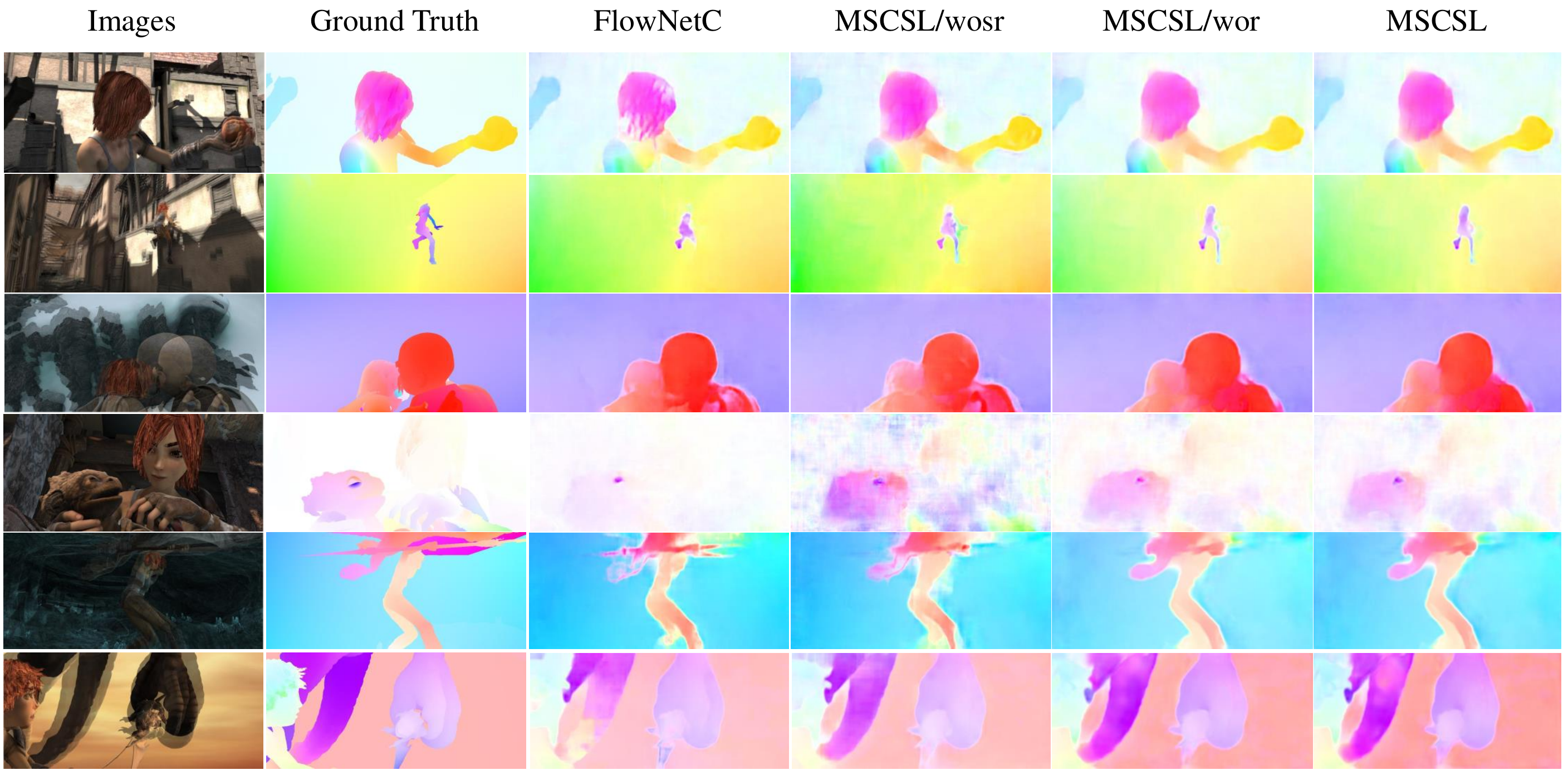}
\caption{Examples of optical flow estimation using FlowNetC, MSCSL/wosr, MSCSL/wor and MSCSL on the MPI Sintel dataset (Clean version). Note that our proposed methods perform well in both small displacement and large displacement.}
\label{sintel}
\end{figure}

\subsection{Implementation Details}
\subsubsection{Network Architecture}

\noindent In this part, we introduce the network architecture briefly. We use $7\times{7}$ convolutional kernel for the first convolutional layer and $5\times{5}$ for the second and third convolutional layers. Then we use max-pooling and convolutional operations to obtain multi-scale representations, as illustrated in Fig.~\ref{architecture}. The correlation layer is the same as that proposed in~\cite{dosovitskiy2015flownet}, and the $d_l$ are set to $5,5,10,10$ from top to bottom (or from coarse to fine). And then we employ $3\times{3}$ kernel and $4\times{4}$ kernel for the other convolutional layers and deconvolutional layers respectively.
\subsubsection{Data Augmentation}

\noindent To avoid overfitting and improve the generalization of network, we employ the data augmentation strategy for the training by performing random online transformations, including scaling, rotation, translation, as well as additive Gaussian noise, contrast, multiplicative color changes to the RGB channels per image, gamma and additive brightness.
\subsubsection{Training Details}

\noindent We implement our architecture using Caffe~\cite{jia2014caffe} and use an NVIDIA TITAN X GPU to train the network. To verify our proposed framework, we conduct three comparison experiments, (1) MSCSL/wosr, this experiment does not contain both the proposed Spatial Conv-GRU and reconstruction loss, and use the refinement network proposed in~\cite{dosovitskiy2015flownet} to predict dense optical flow, (2) MSCSL/wor, this experiment employs the Spatial Conv-GRU, which can be implemented by unfolding the recurrent model in the prototxt file, to encode the correspondence maps for dense optical flow estimation and demonstrates the effectiveness of the Spatial Conv-GRU in comparison to MSCL/wosr, (3) MSCSL, this experiment contains all parts (Spatial Conv-GRU and reconstruction loss) aforementioned.

In the MSCSL/wosr and MSCSL/wor, we train the networks on Flying Chairs training dataset using Adam optimization with $\beta_1=0.9$ and $\beta_2=0.999$. To tackle the gradients explosion, we adopt the same strategy as proposed in~\cite{dosovitskiy2015flownet}. Specifically, we firstly use a learning rate of $1e-6$ for the first $10$k iterations with a batch size of $8$ pairs. After that, we increase the learning rate to $1e-4$ for the following $300$k iterations, and then divide it by $2$ every $100$k iterations. We terminate the training after $600$k iterations (about $116$ hours).

In the MSCSL, we firstly train the MSCSL/wor for $500$k iterations using the training strategy above. After that, we add the reconstruction loss with the balance parameter $\lambda=0.005$. And then we fine-tune the network for $100$k iterations with a fixed learning of $1.25e-5$.

After training the three networks on Flying Chairs training dataset respectively, we fine-tune the networks on the MPI Sintel training dataset for tens of thousands of iterations with a fixed learning rate of $1e-6$ until the networks converge. Specifically, we fine-tune the networks on the Clean version and Final version together with $1,816$ for training and $266$ for validation. Since the KITTI 2012 dataset and Middlebury dataset are small and only contain sparse ground truth, we do not conduct fine-tuning on these two datasets.

\subsection{Comparison to State-of-the-Art}
In this section, we compare our proposed methods to recent state-of-the-art approaches, including traditional methods, such as EpicFlow~\cite{revaud2015epicflow}, DeepFlow~\cite{weinzaepfel2013deepflow}, FlowFields~\cite{bailer2015flow}, EPPM~\cite{bao2014fast}, LDOF~\cite{brox2011large}, DenseFlow~\cite{yang2015dense}, and deep learning based methods, such as FlowNetS~\cite{dosovitskiy2015flownet}, FlowNetC~\cite{dosovitskiy2015flownet}, SPyNet~\cite{ranjan2016optical}.
Table~\ref{epe} shows the performance comparison between our proposed methods and the state-of-the-art using average endpoint errors (EPE). We mainly focus on the deep learning based methods, so we only compare our proposed methods with the learning-based frameworks such as FlowNet and SpyNet.

\begin{figure}[t]
\centering
\includegraphics[scale=0.45]{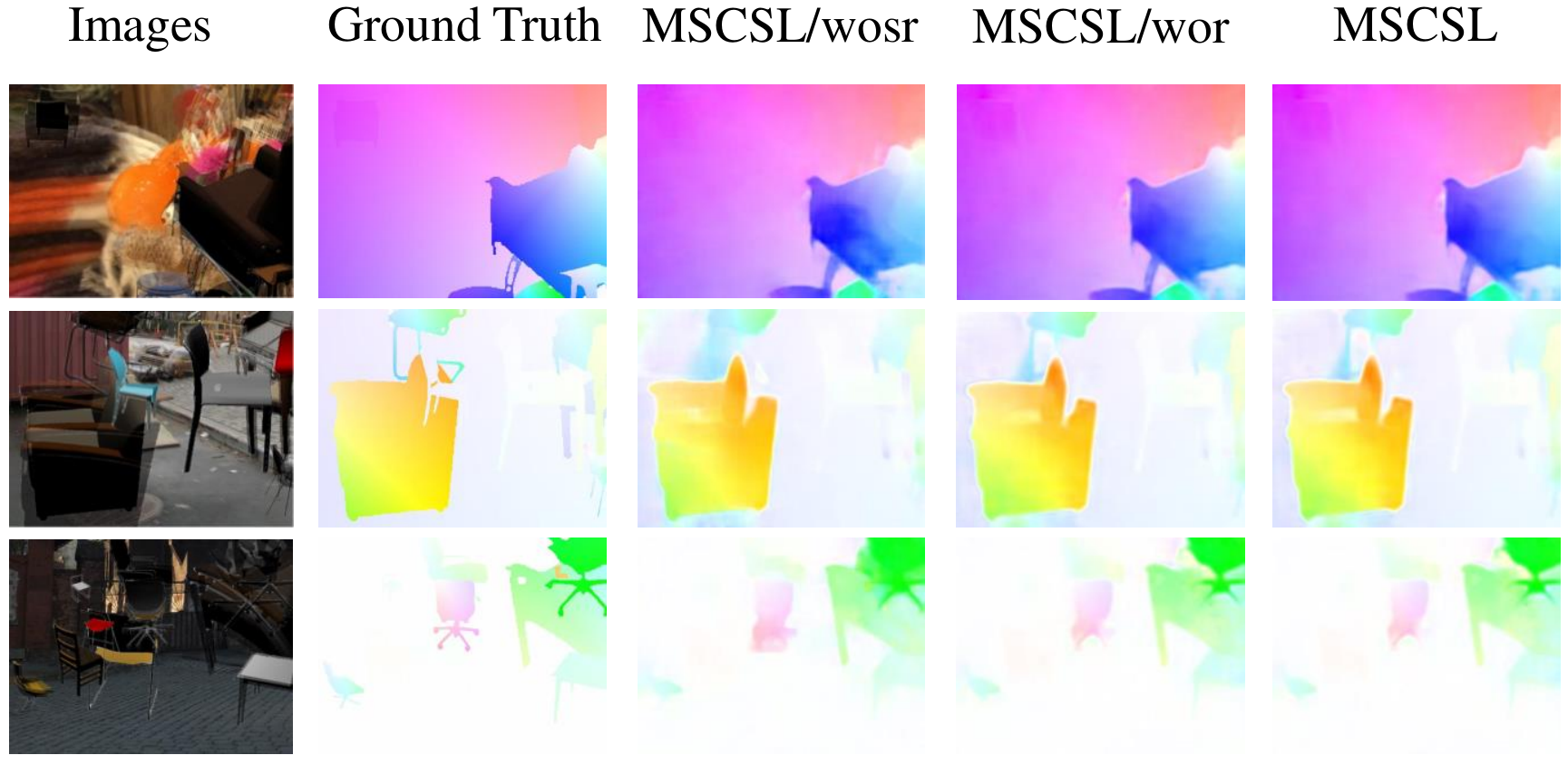}
\caption{Examples of optical flow prediction on the Flying Chairs dataset. Comparison to MSCSL/wosr, the results of MSCSL/wor and MSCSL are more smooth and finer.}
\label{fly}
\end{figure}

\noindent\textbf{Flying Chairs.} For all three comparison experiments, We train our networks on this dataset firstly, and employ MPI Sintel dataset to fine-tune them further. Table~\ref{epe} shows that MSCSL outperforms the other comparison experiments, MSCSL/wosr and MSCSL/wor. Furthermore, our proposed methods achieve better performance comparable with the state-of-the-art methods. After fine-tuning, in most cases most learning based methods suffer from performance decay, this is mostly because of the disparity between Flying Chairs and MPI Sintel dataset. Some visual estimation results on this dataset are shown in Fig.~\ref{fly}.

\noindent\textbf{MPI Sintel.} After the training on Flying Chairs firstly, we fine-tune the trained models on this dataset. The models trained on Flying Chairs are evaluated on the training dataset. The results shown in Table~\ref{epe} demonstrate MSCSL's and MSCSL/sor's better ability to generalize than MSCSL/wosr's and other learning based approaches'. To further verify our proposed methods, we compare our methods with FlowNetS, FlownetC and SPyNet on MPI Sintel test dataset for different velocities and distances from motion
boundaries, as described in Table~\ref{epesintel}. As shown in Table~\ref{epe} and Table~\ref{epesintel}, our proposed methods perform better than other deep learning based methods. However, in the regions with velocities larger than $40$ pixels (smaller than $10$ pixels), the proposed methods are less accurate than FlowNetC (SpyNet). Some visual results are shown in Fig.~\ref{sintel}.

\noindent\textbf{KITTI 2012 and Middlebury.} These two datasets are too small, so we do not fine-tune the models on these datasets. We evaluate the trained models on KITTI 2010 training dataset, KITTI 2012 test dataset and Middlebury training dataset respectively. Table~\ref{epe} shows that our proposed methods outperform other deep learning based approaches remarkably on the KITTI 2012 dataset (including training set and test set). However, in most cases, on Middlebury training dataset, mainly containing small displacements, our proposed methods do not perform well, comparison to SPyNet.

\noindent\textbf{Analysis.}
The results of our framework are more smooth and fine-grained. Specifically, our framework is capable of capturing the motion information of fine-grained object parts, as well as preserving edge information. Meanwhile, our Spatial Conv-GRU can suppress the noises in the results of model without it. All these insights can be observed in Fig.~\ref{fly} and Fig.~\ref{sintel}. However, our proposed frameworks are incapable of effectively capturing the correspondence structure and unstable in regions where the texture is uniform (e.g., on Middlebury dataset).

\noindent\textbf{Timings.}
In Table~\ref{epe}, we show the per-frame runtimes of different approaches. Traditional methods are often implemented on a single CPU, while deep learning based methods tend to run on GPU. Therefore, we only compare the runtimes with FlowNetS, FlowNetC and SPyNet. The results in Table~\ref{epe} demonstrate that our proposed methods (run on NVIDIA TITAN X GPU) improve the accuracy with a comparable speed against the state-of-the-art.

\section{Conclusion}
In this paper, we propose a novel end-to-end multi-scale correspondence structure learning based on deep learning for optical flow estimation. The proposed MSCSL learns the correspondence structure and models the multi-scale dependency in a unified end-to-end deep learning framework. Our model outperforms the state-of-the-art approaches based on deep learning by a considerable computing efficiency. The experimental results on several datasets demonstrate the effectiveness of our proposed framework.

\section*{Acknowledgements}
This work was
supported in part by
the National Natural Science Foundation of China
under Grant U1509206 and Grant 61472353, in part by the
Alibaba-Zhejiang University Joint Institute of Frontier Technologies.

\bibliographystyle{acm}
\bibliography{ijcai17}

\begin{thebibliography}{10}

\bibitem{ahmadi2016unsupervised}
{\sc Ahmadi, A., and Patras, I.}
\newblock Unsupervised convolutional neural networks for motion estimation.
\newblock In {\em Image Processing (ICIP), 2016 IEEE International Conference
  on\/} (2016), IEEE, pp.~1629--1633.

\bibitem{bailer2015flow}
{\sc Bailer, C., Taetz, B., and Stricker, D.}
\newblock Flow fields: Dense correspondence fields for highly accurate large
  displacement optical flow estimation.
\newblock In {\em Proceedings of the IEEE International Conference on Computer
  Vision\/} (2015), pp.~4015--4023.

\bibitem{baker2011database}
{\sc Baker, S., Scharstein, D., Lewis, J., Roth, S., Black, M.~J., and
  Szeliski, R.}
\newblock A database and evaluation methodology for optical flow.
\newblock {\em International Journal of Computer Vision 92}, 1 (2011), 1--31.

\bibitem{ballas2015delving}
{\sc Ballas, N., Yao, L., Pal, C., and Courville, A.}
\newblock Delving deeper into convolutional networks for learning video
  representations.
\newblock {\em arXiv preprint arXiv:1511.06432\/} (2015).

\bibitem{bao2014fast}
{\sc Bao, L., Yang, Q., and Jin, H.}
\newblock Fast edge-preserving patchmatch for large displacement optical flow.
\newblock In {\em Proceedings of the IEEE Conference on Computer Vision and
  Pattern Recognition\/} (2014), pp.~3534--3541.

\bibitem{brox2011large}
{\sc Brox, T., and Malik, J.}
\newblock Large displacement optical flow: descriptor matching in variational
  motion estimation.
\newblock {\em IEEE transactions on pattern analysis and machine intelligence
  33}, 3 (2011), 500--513.

\bibitem{butler2012naturalistic}
{\sc Butler, D.~J., Wulff, J., Stanley, G.~B., and Black, M.~J.}
\newblock A naturalistic open source movie for optical flow evaluation.
\newblock In {\em European Conference on Computer Vision\/} (2012), Springer,
  pp.~611--625.

\bibitem{dosovitskiy2015flownet}
{\sc Dosovitskiy, A., Fischery, P., Ilg, E., Hazirbas, C., Golkov, V., van~der
  Smagt, P., Cremers, D., Brox, T., et~al.}
\newblock Flownet: Learning optical flow with convolutional networks.
\newblock In {\em 2015 IEEE International Conference on Computer Vision
  (ICCV)\/} (2015), IEEE, pp.~2758--2766.

\bibitem{geiger2012we}
{\sc Geiger, A., Lenz, P., and Urtasun, R.}
\newblock Are we ready for autonomous driving? the kitti vision benchmark
  suite.
\newblock In {\em Computer Vision and Pattern Recognition (CVPR), 2012 IEEE
  Conference on\/} (2012), IEEE, pp.~3354--3361.

\bibitem{jaderberg2015spatial}
{\sc Jaderberg, M., Simonyan, K., Zisserman, A., et~al.}
\newblock Spatial transformer networks.
\newblock In {\em Advances in Neural Information Processing Systems\/} (2015),
  pp.~2017--2025.

\bibitem{jason2016back}
{\sc Jason, J.~Y., Harley, A.~W., and Derpanis, K.~G.}
\newblock Back to basics: Unsupervised learning of optical flow via brightness
  constancy and motion smoothness.
\newblock In {\em Computer Vision--ECCV 2016 Workshops\/} (2016), Springer,
  pp.~3--10.

\bibitem{jia2014caffe}
{\sc Jia, Y., Shelhamer, E., Donahue, J., Karayev, S., Long, J., Girshick, R.,
  Guadarrama, S., and Darrell, T.}
\newblock Caffe: Convolutional architecture for fast feature embedding.
\newblock In {\em Proceedings of the 22nd ACM international conference on
  Multimedia\/} (2014), ACM, pp.~675--678.

\bibitem{ranjan2016optical}
{\sc Ranjan, A., and Black, M.~J.}
\newblock Optical flow estimation using a spatial pyramid network.
\newblock {\em arXiv preprint arXiv:1611.00850\/} (2016).

\bibitem{revaud2015epicflow}
{\sc Revaud, J., Weinzaepfel, P., Harchaoui, Z., and Schmid, C.}
\newblock Epicflow: Edge-preserving interpolation of correspondences for
  optical flow.
\newblock In {\em Proceedings of the IEEE Conference on Computer Vision and
  Pattern Recognition\/} (2015), pp.~1164--1172.

\bibitem{siam2016convolutional}
{\sc Siam, M., Valipour, S., Jagersand, M., and Ray, N.}
\newblock Convolutional gated recurrent networks for video segmentation.
\newblock {\em arXiv preprint arXiv:1611.05435\/} (2016).

\bibitem{weinzaepfel2013deepflow}
{\sc Weinzaepfel, P., Revaud, J., Harchaoui, Z., and Schmid, C.}
\newblock Deepflow: Large displacement optical flow with deep matching.
\newblock In {\em Proceedings of the IEEE International Conference on Computer
  Vision\/} (2013), pp.~1385--1392.

\bibitem{yang2015dense}
{\sc Yang, J., and Li, H.}
\newblock Dense, accurate optical flow estimation with piecewise parametric
  model.
\newblock In {\em Proceedings of the IEEE Conference on Computer Vision and
  Pattern Recognition\/} (2015), pp.~1019--1027.

\bibitem{zhu2017guided}
{\sc Zhu, Y., Lan, Z., Newsam, S., and Hauptmann, A.~G.}
\newblock Guided optical flow learning.
\newblock {\em arXiv preprint arXiv:1702.02295\/} (2017).

\end{thebibliography}

\end{document}